\pdfoutput=1

\documentclass[11pt]{article}

\usepackage{EMNLP2022}

\usepackage{times}
\usepackage{latexsym}

\usepackage[T1]{fontenc}

\usepackage[utf8]{inputenc}
\usepackage{microtype}
\usepackage{inconsolata}

\usepackage{hyperref}       
\usepackage{url}            
\usepackage{booktabs}       
\usepackage{amsfonts}       
\usepackage{nicefrac}       
\usepackage{xcolor}         

\usepackage{xspace}
\newcommand{\ours}{MemSizer\xspace}

\def\vphi{{\boldsymbol{\phi}}}

\def\vb{{\mathbf{b}}}

\def\vk{{\mathbf{k}}}

\def\vq{{\mathbf{q}}}

\def\vv{{\mathbf{v}}}

\def\vx{{\mathbf{x}}}

\def\valpha{{\mathbf{\alpha}}}

\def\mI{{\mathbf{I}}}

\def\mK{{\mathbf{K}}}

\def\mQ{{\mathbf{Q}}}

\def\mV{{\mathbf{V}}}
\def\mW{{\mathbf{W}}}
\def\mX{{\mathbf{X}}}

\def\mPhi{{\mathbf{\Phi}}}

\def\gO{{\mathcal{O}}}

\usepackage{tabulary}
\usepackage{amsmath}
\usepackage{amssymb}
\usepackage{MnSymbol}%
\usepackage{wasysym}%
\usepackage{tikz}
\usepackage{xcolor}
\usepackage{color, colortbl}
\usepackage[normalem]{ulem}
\usepackage{url}
\usepackage{multirow}
\usepackage{float}
\usepackage{paralist}
\usepackage{comment}

\newcommand{\out}{{\mathrm{out}}}

\newcommand{\elu}{{\operatorname{elu}}}

\usepackage{mathtools}
\usepackage{arydshln}
\usepackage{subcaption}
\usepackage{enumitem}
\setlist{noitemsep,topsep=0pt,parsep=0pt,partopsep=0pt, leftmargin=12pt}

\title{Linearizing Transformer with Key-Value Memory}

\author{
Yizhe Zhang\thanks{~~Equal contribution.} \\
    Meta AI \\
    {\texttt yizhezhang@fb.com} \\
\And
    Deng Cai\footnotemark[1] \\
    The Chinese University of Hong Kong \\
    {\texttt thisisjcykcd@gmail.com}\\
}

\begin{document}
\maketitle
\begin{abstract}
Efficient transformer variants with linear time complexity have been developed to mitigate the quadratic computational overhead of the vanilla transformer. Among them are low-rank projection methods such as Linformer and kernel-based Transformers. Despite their unique merits, they usually suffer from a performance drop comparing with the vanilla transformer on many sequence generation tasks, and often fail to obtain computation gain when the generation is short. 
We propose \ours, an approach towards closing the performance gap while improving the efficiency even with short generation. It projects the source sequences into lower dimension representations like Linformer, while enjoying efficient recurrent-style incremental computation similar to kernel-based transformers.
This yields linear computation time and constant memory complexity at inference time. 
\ours also employs a lightweight multi-head mechanism which renders the computation as light as a single-head model.  
We demonstrate that \ours provides an improved balance between efficiency and accuracy over the vanilla transformer and other efficient transformer variants in three typical sequence generation tasks, including machine translation, abstractive text summarization, and language modeling.
\end{abstract}

\section{Introduction}
Transformer \citep{Vaswani2017AttentionIA} has become the \textit{de facto} standard for almost all NLP tasks across the board. At the core of the vanilla transformer is the attention mechanism that captures the interactions between feature vectors at different positions in a sequence.
Despite its great success, the vanilla transformer models are typically computationally expensive as the computation of the attention mechanism scales quadratically with the sequence length. This bottleneck limits the efficient deployment of large-scale pre-trained models, such as GPT-3 \citep{gpt3}, Image Transformer \citep{imagetransformer}, Codex \citep{chen2021evaluating} and DALL-E \citep{DALLE}. Training and deploying such gigantic transformer models can be prohibitively difficult for scenarios with limited resource budgets and may result in huge energy consumption and greenhouse gas emission \citep{strubell2019energy, schwartz2020green}. 

A number of transformer variants have been proposed to reduce the computational overhead \citep{Tay2020EfficientTA}. One family of methods leverages low-rank projections to reduce the number of pair-wise interactions (\textit{i.e.}, the size of attention matrices) \citep{Wang2020LinformerSW,nystom,synthesizer}. These methods first project the input sequence into a low-resolution representation. For example, \citet{Wang2020LinformerSW} project the length dimension to a fixed feature dimension. Nevertheless, these methods have difficulties modeling variable-length sequences and autoregressive (causal) attention, impeding their applications in sequence generation tasks. Recent works propose to approximate the softmax attention through kernelization \citep{katharopoulos-et-al-2020,RFA,performer,kasai2021finetuning}. For sequence generation tasks, these works can cache computation in a \textit{recurrent} manner, leading to constant memory complexity in sequence length during inference. Despite the improved efficiency in long-form generation, the computation gain of these kernel-based approaches vanishes when the generation is as short as a typical sentence length. Additionally, they usually suffer from a performance loss when training from scratch \citep{kasai2021finetuning}. 

In this work, we propose an approach called \ours, an efficient transformer variant which follows the paradigm of low-rank projections while enjoying memory-efficient recurrent-style generation as the kernel-based transformers. Concretely, we develop a key-value memory layer \citep{sukhbaatar2015end} to substitute the multi-head attention layer in the vanilla transformer. We pack the information in the source sequence into a fixed-sized set of memory values in a length-dynamic manner, and use input-independent parametric matrices as the memory keys.  
In this way, we emphasize more on modeling the values and significantly simplify the design of keys.
This unbalanced design of keys and values further enables us to suppress the multi-head computation to be as fast as with single head. 

\ours is conceptually simple yet can handle variable-length sequences and causal attention for generation thanks to the length-dynamic projection. With the unbalanced memory layer and dynamic projection, \ours enjoys linear time complexity and constant memory complexity. 
Our experiments in three typical sequence generation tasks (machine translation, abstractive text summarization, and language modeling) show that the proposed method achieves comparable or better performance to state-of-the-art linear recurrent transformers, with more substantial reductions in inference latency, memory consumption, and model size. The advantages are more prominent with longer input lengths. 
In some tasks, the proposed \ours can maintain or even improve the performance of the vanilla transformer, offering an appealing alternative for sequence generation tasks. 


\section{Preliminaries}
\subsection{Key-Value Memory Networks}
\label{mem}
We first review the general ideas of memory networks \citep{graves2014neural,sukhbaatar2015end}. In a nutshell, given a set of \textit{source} vectors $\mX^{s} = \{ \vx^{s}_{i}\}_{i=1}^M$, a basic key-value memory network first projects the entire set into memory key vectors $\mK \in \mathbb{R}^{M \times h}$ and value vectors $\mV \in \mathbb{R}^{M \times h}$ respectively. A \textit{target} vector $\vx^{t}$ for querying the key-value memories will also be embedded as $\vq \in \mathbb{R}^h$ which shares the same embedding space of $\mK$. This is followed by computing a probability vector over the key vectors according to the inner product similarity: 
\begin{align}
  \valpha = f (\vq\mK^T), \label{eq:xk} 
\end{align}%
where $f$ denotes an activation function. A typical choice for $f$ is the softmax function. The output vector $\vx^{\out}$, which can be used for final prediction or next layer's input, is simply summarizing over the value vectors according to their probabilities:
\begin{align}
  \vx^{\out}  = \valpha \mV. \label{eq:vh} 
\end{align}%



\subsection{Transformer}
\label{trans}
\paragraph{Architecture}
The vanilla transformer architecture consists of multi-head attention, feedforward layers, and layer normalization modules \citep{Vaswani2017AttentionIA}. 
The multi-head attention module (referred to as standard attention, or SA, throughout this paper) plays a core role in a vanilla transformer. SA takes input as sequences of \textit{source} and \textit{target} vectors.
The source vectors are used to produce \textit{key} and \textit{value} features, while the target vectors are mapped to \textit{query} vectors. We denote the source and target vectors by $\mX^{s} \in \mathbb{R}^{M \times d}$ and $\mX^{t} \in \mathbb{R}^{N \times d}$, where $d$ is the model dimensionality. The input vectors for each head are first mapped to $h$-dimensional \textit{query}, \textit{key}, and \textit{value} features by learned affine transformations with $\mW_\ast \in \mathbb{R}^{d \times h}$ and $\vb_{\ast} \in \mathbb{R}^{h}$:%
\begin{align}
\mQ &=  \mX^{t}\mW_{q} + \vb_q, \quad
\mK =  \mX^{s}\mW_{k}  +\vb_k, \label{eq:q}\\
\mV &= \mX^{s}\mW_{v} + \vb_v. 
\label{eq:kv}
\end{align}
The attention is achieved by computing the normalized similarities of query vector and key vectors:
\begin{equation}
\valpha = \text{softmax}({\frac{\mQ\mK^{T}}{\sqrt{h}}}).
\label{eq:attn}
\end{equation}
The attention weights $\valpha$ are then used to calculate a weighted average of the value vectors as in eq (\ref{eq:vh}). It is generally assumed there are $r$ attention heads of $h$-dimensional such that $d = hr$. SA performs above procedure for each of the $r$ heads in parallel and concatenates $r$ output vectors to get the final $d$-dimensional vector:\footnote{The layer normalization \citep{Ba2016LayerN} and residual connection \citep{resnet} steps are suppressed for brevity.}  
\begin{align}
 \mX^{\out} = [\mX^{\out}_{(1)}, \ldots, \mX^{\out}_{(r)}]W_o+b_o,
\end{align}%
where $W_o \in \mathbb{R}^{d \times d}$ and $b_o \in \mathbb{R}^{d}$ are the output projection weights.
\paragraph{Time Complexity}
The computation in a transformer can be divided into three stages:

$(i)$
\textsc{Feature Mapping}: The time complexity of the computation of $\mQ$,  $\mK$, and $\mV$ for all $r$ heads (Eq.\ (\ref{eq:q}-\ref{eq:kv})) is $\gO(Nd^2)$, $\gO(Md^2)$, and $\gO(Md^2)$, respectively.

$(ii)$
\textsc{Attention}: The time complexity of the computation of attention matrices for all $r$ heads (Eq.\ (\ref{eq:attn})) is $\gO(MNd)$, which scales quadratically in sequence length ($M$, $N$).

$(iii)$
\textsc{Projection}: The time complexity of projecting the concatenated $\vx^{\out}$ from $r$ heads back to $d$-dimensional vector is $\gO(Nd^2)$.

Taking all three parts together, a SA module scales at $\gO(MNd + Md^2 +Nd^2)$. When sequence length is long ($M, N \gg d$), $\gO(MNd)$ will dominate the computation.

\paragraph{Memory Complexity}
At every generation step, query, key, and value vectors consume space complexity of $\gO(d)$, $\gO(Md)$, and $\gO(Md)$, respectively.
Every step's attention weight (Eq.~(\ref{eq:attn})) attends across $M$ source positions, consuming $\gO(Mr)$ space.

\section{\ours: A Different Perspective of Attention Mechanism}
\label{sec:memsizer}
As discussed in Section \ref{trans}, the SA in the vanilla transformer can be perceived as an instantiation of the key-value memory network in Section \ref{mem}, where the memory key $\mK$ and value $\mV$ are point-wise projections of the source $\mX^{s}$.
In this work, we replace the SA module with a different memory mechanism which achieves recurrent inference computation thus linear complexity. Our memory mechanism comes with a different specification of query, key and value in SA. Specifically, following Eq. (\ref{eq:xk}-\ref{eq:vh}), we specify the key-value memory layer as
\begin{align}
\mQ &= \mX^{t}, \quad \mK = \mPhi,  \label{eq:k}\\
\mV & = \text{LN}(\mW_l (\mX^{s})^T)  \text{LN}(\mX^{s} \mW_r).
\label{eq:v}
\end{align}

\paragraph{Unbalanced Key-value Memory Mechanism}
The key-value memory layer in \ours contains $k$ memory slots. Inspired by \citet{synthesizer},
which demonstrates that query-key attention can be significantly simplified in the vanilla transformer, the key matrix $\mPhi \in \mathbb{R}^{k \times d}$ in Eq. (\ref{eq:k}) is a learnable parametric matrix, which is input-independent and shared across different instances.
The value matrix in \ours, likewise, contains $k$ memory value vectors of $d$ dimension. It summarizes the source information into a fixed-sized space $\mathbb{R}^{k\times d}$ regardless of the source length $M$. 
Compared with the vanilla Transformer which treats keys and values equally, this unbalanced key-value mechanism emphasizes learning better input-dependent values to match with input-independent keys.

\paragraph{Values Matrix via Dynamic-length Projection}
To pack the source information into the $\mathbb{R}^{k\times d}$ value matrix, Linformer \citep{Wang2020LinformerSW} uses a low-rank projection. However, performing the low-rank projection would require the input sequence length $M$ to be preset before training, making Linformer difficult to be applied to scenarios with dynamic input length and generation tasks.
To solve this issue, we apply a linear kernel $({\mX}^T\mX)$ to the source input $\mX^{s}$ to cancel out the length dimension $M$, so that $M$ is not required to be preset. The value matrix essentially captures the second moment (covariance) information from the source $\mX^{s}$ \citep{el2021xcit,zhu2021long}. We use two adaptor projection matrices $\mW_l \in \mathbb{R}^{k \times d}$ and $\mW_r \in \mathbb{R}^{d \times d}$ to project source information into $k$ global token-independent memory value vectors\footnote{Note that the inclusion of $\mW_r$ does not affect the dimensionality of $\mV$. However, in our experiments removing the $\mW_r$ will harm the performance.}. 
The value matrix $\mV$ is formulated in Eq. (\ref{eq:v}), where
LN$(\cdot)$ denotes the layer normalization \citep{Ba2016LayerN}, which makes the training robust in our experiments. To control the magnitude of $V$ across variable-length input sequences, we multiply the $\mV$ by a scaling factor of $1/{\sqrt{M}}$, which resembles the rescaling rationale from SA in Eq. (\ref{eq:attn}). 

\paragraph{Lightweight Multi-head Computation}
The model can be made more expressive with multi-head specification, where we share $\mV$ across $r$ different heads but use a distinct $\mK$ for each head. Following \citet{lample2019large}, the outputs from each head are simply aggregated through mean-pooling. Specifically, 
\begin{align}
 \mX^{\out} = 1/r \cdot \sum_{i=1}^r \mX^{\out}_{(i)},
\end{align}%
where $\mX^{\out}_{(i)}$ is the output from $i$-th head. The final output $\mX$ has dimension $d$, therefore the output projection layer in the vanilla transformer is \textit{no longer} needed. 

In \ours, the above multi-head computation is \textit{negligible}, as it can be done by first averaging the attention weights $\valpha$ in Eq. (\ref{eq:xk}) from different heads into $\bar{\valpha}$, followed by as if performing single-head attention using $\bar{\valpha}$. The overall computation is as lightweight as a single-head model. 

\paragraph{Recurrent Computation for Memory-efficient Generation}
Similar to previous kernel-based transformers \citep{kasai2021finetuning, RFA}, generation computation in \ours can be rolled out as a recurrent procedure. At each generation step $i$, define $\mV_i$ as the recurrent states ~\citep{katharopoulos-et-al-2020}:
\begin{align}
     \mV_i = \sum_{j=1}^i \text{LN}(\mW_l (\vx^{s}_j)^T)  \text{LN}(\vx^{s}_j \mW_r),
\end{align}
where $\vx^{s}_j$ is the $j$-th row of $\mX^{s}$, $\mV_i$ can be perceived as a rolling-sum matrix: 
\begin{align}
     \mV_i = \mV_{i-1} + \text{LN}(\mW_l (\vx^{s}_i)^T)  \text{LN}(\vx^{s}_i \mW_r).
\end{align}

Consequently, the output $\vx^{\out}_i$ can be computed in an incremental manner from cached recurrent matrix $V_{i-1}$. This avoids quadratic computation overhead in input sequence length.
\paragraph{Time Complexity}
We break down the time complexity of each step in \ours.
\ours proceeds over two stages which correspond to the first two stages of SA. The last output projection stage in SA does not exist in \ours.

$(i)$
\textsc{Memory Projection}: To obtain the value matrix $\mV$ (Eq. (\ref{eq:v}), shared over heads), we first compute $\mW_l (\mX^{s})^T$ and $\mX^{s} \mW_r $, of which the time complexity is $\gO( Mdk )$ and $\gO( Md^2)$, respectively. The product of $\mW_l (\mX^{s})^T$ and $\mX^{s} \mW_r $ further takes $\gO(Mdk)$. In total, $\gO(Md^2 + Mdk)$.

$(ii)$
\textsc{Attention}: The attention computation (Eq. (\ref{eq:xk})) is computed with $\gO (Ndk)$.

Taking both parts together, the attention mechanism in \ours scales with $\gO(Mdk+ Md^2 + Ndk)$. Compared to $\gO(MNd + Md^2 +Nd^2)$ of SA, we see that if the number of memory slots $k$ is much smaller than sequence lengths ($k<<M, N$), the change of time complexity from $\gO(MNd)$ to $\gO(Mdk)+\gO(Ndk)$ brings a substantial speedup.
\paragraph{Memory Complexity}
\ours only needs to store the value matrix $\mV$, and thus its space complexity is $\gO(dk)$, constant in sequence length. 
This implies a reduction in memory footprints when $k<<M$, compared to SA's $\gO(Md)$.

\paragraph{Comparison with Other Transformers}
Compared with the vanilla transformer, 
each memory slot value $\vv_{j\in \{1, \cdots, k\}}$ summarizes a \textit{global position-agnostic} feature of the source context $\mX^{s}$. \ours enjoys linear time complexity as Linformer and additionally possesses the advantage of recurrent-style sequence generation as kernel-based transformers. A detailed comparison among \ours, the vanilla transformer and other efficient transformers is in the Appendix~\ref{app:comp} (Table~\ref{comp}). 

\section{Experiments}
\label{sec:experiments}
We present extensive experiments on three typical sequence generation tasks in NLP, including machine translation, abstractive text summarization, and language modeling.
\subsection{Baselines}
We compare \ours with previous transformer variants with \textit{linear} time complexity and \textit{constant} memory complexity in input sequence length, which limits the comparison to kernelization approaches \citep{katharopoulos-et-al-2020, RFA, performer, kasai2021finetuning}. Linformer assumes a fixed sequence length. This makes Linformer suit well with understanding tasks but difficult to be applied to generation tasks, as generation tasks typically assume variable generation length and autoregressive (causal) attention Likewise, Synthesizer needs to specify the maximum input length and thus does not suit well tasks with variable generation lengths. Thus Linformer and Synthesizer are excluded from the comparison.
The compared methods correspond to three different feature maps $\vphi$: \textbf{ELU} ($\vphi\left(\vx\right) = \elu \left(\vx\right) + 1$, \citealp{katharopoulos-et-al-2020}); \textbf{RFA} (random feature approximation with softmax temperature reparameterization, \citealp{RFA, katharopoulos-et-al-2020}); \textbf{T2R} (trainable random feature). \textbf{Performer} \citep{performer} employs a similar random approximation to RFA. We omitted it from the comparison as it diverges during training in our experiments.
All models are randomly initialized via Xavier initialization \citep{glorot2010understanding}.
\begin{table*}[h]
\centering
\small
\addtolength{\tabcolsep}{-0.0pt}  
\begin{tabular}{@{} lcccccccc @{}}
\toprule
\textbf{Model} &  \multicolumn{2}{c}{$k$ (cross, causal)} &   
\multicolumn{1}{c}{En-De} & En-Fr & Zh-En & Speed & Memory  & Model size\\ 
 \hline
ELU \ & 64 & 64 & 28.4 & * & 23.4 & 4605.6 & 9.842G &   209M\\
RFA \ & 32 & 4 & 28.1 & 41.7 & 23.4 & 3771.6  & 4.058G & 210M \\
T2R \ &32 &  4 & 27.5 & 39.8 & 23.1 & 5408.4  & 4.057G & 210M\\
\hdashline
\ours  & 32 & 4& \textbf{28.4} & \textbf{42.4} & \textbf{24.5} & \textbf{7476.3} & \textbf{3.896}G  & \textbf{176M} \\
\hline
Transformer  & -- &-- &  28.9 & 42.2 & 24.2 & 5506.5 & 5.537G & 209M \\
\bottomrule
\end{tabular}
\caption{Machine translation test results on MT datasets. The results for baselines are from \citet{kasai2021finetuning}. The vanilla transformer is implemented following \citet{Vaswani2017AttentionIA}. (\citet{Vaswani2017AttentionIA} reports BLEU$=28.4$ for En-De and 41.8 for En-Fr, which is worse than this implementation). ``*'' indicates divergence during training. 
The inference speed (Speed) measured in the number of tokens per second, peak memory usage (Memory), and model size are benchmarked on En-De translation task. }
\label{mt_results}
\end{table*}
\subsection{Machine Translation}
\paragraph{Setup}We experiment with WMT16 En-De (4.5M train pairs, average target length 29.5 tokens), WMT14 En-Fr (36M, 31.7) and WMT17 Zh-En (20M, 28.5) translation benchmarks \citep{wmt2016-findings}.
We follow the experiment setup, preprocessing and data splits by previous work \citep{kasai2021finetuning}.
Following \citet{Vaswani2017AttentionIA}, we use the large-sized transformer with 6 layers, 16 attention heads, 1024 model dimensions, and 4096 hidden dimensions for both the encoder and decoder.
We apply dropout with $0.3$, weight decay with $0.01$, and label smoothing with $\varepsilon=0.1$.
Following \citet{Ott2018ScalingNM}, we use an increased batch size of approximately 460K tokens by accumulating gradients without updating parameters.
Each model is trained from random initialization for 30K 
(60K for the large En-Fr dataset) 
steps using Adam with a learning rate of $5\cdot10^{-4}$ and $\beta=(0.9, 0.98)$ \citep{Kingma2014AdamAM}. We employ beam search decoding with beam size 5 and length penalty $1.0$ \citep{wu2016google}. 
The checkpoints from the last five epochs are averaged to obtain the final model \citep{Vaswani2017AttentionIA}. 
Following previous works, we use tokenized BLEU \citep{Papineni2001BleuAM} for evaluation.
Our method is applied to both cross and causal attention. 
Following \citet{kasai2021finetuning}, we use memory sizes $k=(32,4)$ for cross and causal attention.

\paragraph{Results} Table \ref{mt_results} presents machine translation results. In general, the kernel-based transformers suffer from additional overhead when the generated sequence is relatively short ($\sim30$ tokens in this task), leading to an incremental speedup compared with the vanilla transformer. ELU has a much larger feature size $k$, leading to increased memory overhead. 
With $\sim17$\% smaller model size, \ours outperforms RFA and T2R while being comparable to ELU, in terms of test BLEU score in En-De. In En-Fr and Zh-En, \ours outperforms all baseline methods including the vanilla transformer. 

As a result of significantly reduced model size, \ours achieves faster generation time and more efficient GPU memory utilization compared to other linear recurrent transformer variants.
\subsection{Abstractive Text Summarization}
\paragraph{Setup} We evaluate on two popular datasets, namely CNN/DailyMail \citep{hermann2015teaching} and XSUM \citep{narayan-etal-2018-dont}. 
We used the standard splits of \citep{nallapati-etal-2016-abstractive} for training, validation, and testing (287,113/13,368/11,490 documents). 
The average lengths of articles and highlights are 766 and 53 respectively. The XSUM dataset \citep{narayan-etal-2018-dont} consists of 227K (204,045/11,332/11,334 for training/validation/testing) BBC articles covering a wide variety of subjects. 
The average lengths of articles and summaries are 431 and 23 respectively. 

We follow \citet{lewis-etal-2020-bart} for data preprocessing and model  configuration. We use the BART-large configuration with 12 layers, 16 attention heads, 1024 model dimensions, and 4096 hidden dimensions for both the encoder and decoder. We apply dropout with $0.1$, weight decay with $0.01$, and label smoothing with $\varepsilon=0.1$. Each model is trained from random initialization for 50K steps using Adam \citep{Kingma2014AdamAM}. We employ beam search decoding with length penalty as in \citet{lewis-etal-2020-bart}. 
We use the standard ROUGE metrics (F1 scores ROUGE-1/2/L) \citep{lin-2004-rouge} for evaluation.
Following the settings in machine translation, we use memory sizes $k=(32,4)$ for cross and causal attention.

\begin{table*}[h]
\centering
\small
\addtolength{\tabcolsep}{-0.0pt}  
\begin{tabular}{@{} lcccccccccc@{}}
\toprule
\multirow{2}{*}{\textbf{Model}} &  \multicolumn{2}{c}{$k$}&   \multicolumn{3}{c}{XSUM}
& \multicolumn{3}{c}{CNN/DailyMail} &\multirow{2}{*}{Speed}&\multirow{2}{*}{Memory}\\ 
\cmidrule(lr){2-3}  \cmidrule(lr){4-6} \cmidrule(lr){7-9} 
&cross&casual&R1 & R2&RL&R1 & R2&RL&&\\
 \hline
Lead-3 &&&16.3 &1.6 &12.0 &40.4  &17.6 &36.7 && \\
RFA \ & 32 & 4 & 28.0 & 9.0 & 22.4 & 35.0 & 10.7& 31.9&323.4&8.6G\\
T2R \ &32 &  4 & 28.6 & 9.3 & 22.8 & 35.8 & 11.2 & 32.7 &358.3&6.2G\\
\hdashline
\ours  & 32 & 4& \textbf{32.3} & \textbf{11.6} & \textbf{25.8} & \textbf{36.3} & \textbf{12.1} & \textbf{33.1} &\textbf{412.3}&\textbf{5.9G}\\
\hline
Transformer  & -- &-- &  31.8 & 11.3 & 25.3 & 39.1 & 15.3 & 35.8 & 338.6 & 23.4G\\
\citet{zhang2020pegasus}  & -- &-- &  30.8 & 10.8 & 24.4 & 38.3 & 15.0 & 35.5 & - & -\\
\bottomrule
\end{tabular}
\caption{Summarization test results on XSUM and CNN/DailyMail datasets. The inference speed (Speed) measured in the number of tokens per second and peak memory usage (Memory) are benchmarked on XSUM dataset. The last row is from \citet{zhang2020pegasus} with the same transformer architecture in our Transformer baseline.}
\label{sum_results}
\end{table*}
\paragraph{Results}
Table \ref{sum_results} presents abstractive text summarization results in ROUGE scores. \ours outperforms RFA and T2R on both datasets in terms of ROUGE scores.\footnote{We omitted the results of ELU because it diverged during training in our experiments.} On the XSUM dataset, \ours even achieves better results than the vanilla transformer while being much faster and memory-efficient. On the CNN/DailyMail dataset, however, there are still considerable performance gaps between \ours and the vanilla transformer. We attribute it to the distinct characteristics of the two datasets. XSUM contains highly abstractive summaries while the summaries in the CNN/DailyMail tend to be more extractive. In fact, the Lead-3 baseline \citep{zhang2020pegasus} outperforms all presented models. We hypothesize that \ours may suffer from the limited capacity of the reduced memory bank for memorizing the exact wordings in the source documents.

Similar to machine translation, the kernel-based transformers suffer from additional overhead when the generated sequence is relatively short ($\sim30$ tokens for summaries), leading to an incremental speedup compared with the vanilla transformer. However, the reduction in peak memory consumption is substantial. This is because the lengthy input documents are packed into a fixed-sized key-value memory bank. Overall, \ours achieves the largest speed-up (22\% speed-up compared to the vanilla transformer) and the smallest memory consumption (75\% reduction compared to the vanilla transformer).

\subsection{Language Modeling}
\paragraph{Setup}
For the first task, we use the WikiText-103 language model (LM) benchmark, which consists of 103M tokens sampled from English Wikipedia \citep{wiki103}.
Following \citet{kasai2021finetuning}, we choose similar hyperparameters to prior work \citep{Baevski2019AdaptiveIR, layerdrop}: 32 layers, 8 heads, 128 head dimensions, 1024 model dimensions, 4096 fully connected dimensions and dropout \citep{dropout} and layer dropout rates of 0.2. We set the memory size $k$ to be 32.
The word embedding and softmax matrices are tied \citep{Press2017UsingTO, Inan2017TyingWV}.
We partition the training data into non-overlapping blocks of 512 contiguous tokens and train the model to autoregressively predict each token \citep{Baevski2019AdaptiveIR}.
Validation and test perplexities are measured by predicting the last 256 words out of the input of 512 consecutive words to avoid evaluating tokens in the beginning with limited context (\textit{early token curse}, \citealp{shortformer}).
We generally follow the optimization method from \citet{Baevski2019AdaptiveIR}, with a slight modification for some hyperparameters including learning rate (we use $10^{-4}$), which shows better convergence. To evaluate the time and memory efficiency of \ours in sequence generation, we generate 256 tokens for each method. 
The batch size is set to be 256.
\begin{table}[t]
\small
\centering
\addtolength{\tabcolsep}{-2.7pt}  
\begin{tabular}{@{} lcccccc @{}}
\toprule

\multirow{2}{*}{\textbf{Model}} &\multirow{2}{*}{$k$} &  \multicolumn{2}{c}{PPL} & \multirow{2}{*}{Speed} & \multirow{2}{*}{Memory}&  \multirow{2}{*}{Model}\\
\cmidrule(lr){3-4} 
&  &dev.  & test &  &  &  Size\\
\hline
ELU  & 128 & 22.0 & 22.8 & 2491 & 6.825G & 449M \\
RFA  & 32 & 20.4 & 21.3 & 2311 & 3.731G & 449M\\
T2R  & 32 & \textbf{20.1} & 20.8 & 2692 & 3.733G & 450M \\
\hdashline
\ours  & 32 & 20.2 &\textbf{20.8}  & \textbf{3165} & \textbf{3.373G} & \textbf{357M} \\
\hline
Transformer  & -- & 17.9 & 18.5 & 1932 & 19.21G & 448M \\
\bottomrule
\end{tabular}
\caption{
WikiText-103 language modeling results in perplexity. The speed is measured for free text generation in the number of tokens per second. The top three rows are implementations from \citet{kasai2021finetuning}. The vanilla transformer is implemented according to \citet{Baevski2019AdaptiveIR}, which reports the test perplexity to be 18.7 (worse than our 18.5 result).
}
\label{lm_results}
\end{table}

	\begin{figure*}[ht!]
		\begin{subfigure}{0.49\textwidth}
		\includegraphics[scale=0.21]{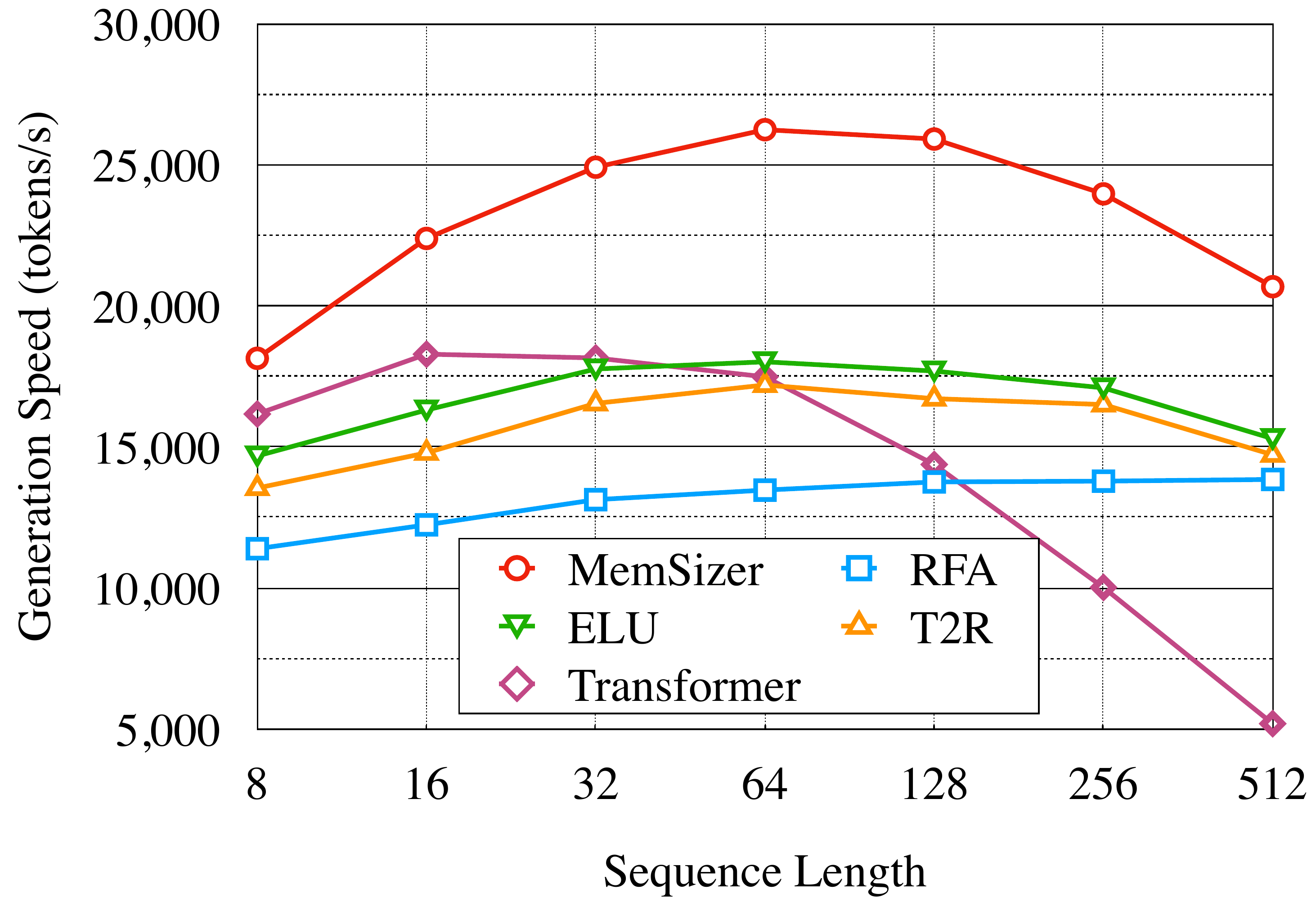}
		\caption{Generation speed.}
		\label{time.len}
	\end{subfigure}
    \begin{subfigure}{0.49\textwidth}
		\includegraphics[scale=0.21]{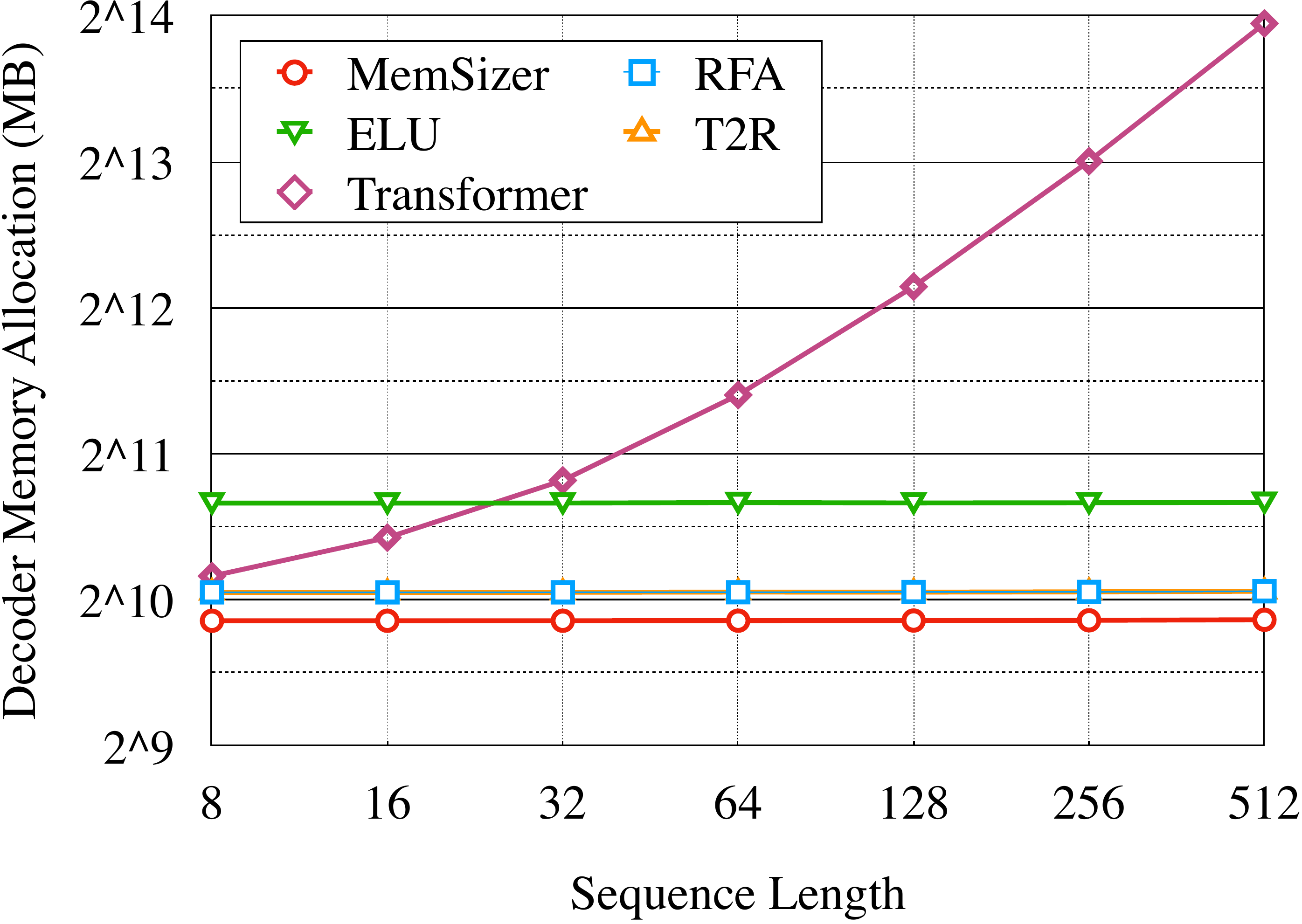}
		\caption{Peak memory consumption. 
		}
		\label{memory.len}
		\end{subfigure}
		\caption{Computational overhead of machine translation (En-Dn) of different sequence lengths.}
	\end{figure*}
	\begin{figure*}[ht!]
		\centering
		\begin{subfigure}{0.49\textwidth}
		\includegraphics[scale=0.21]{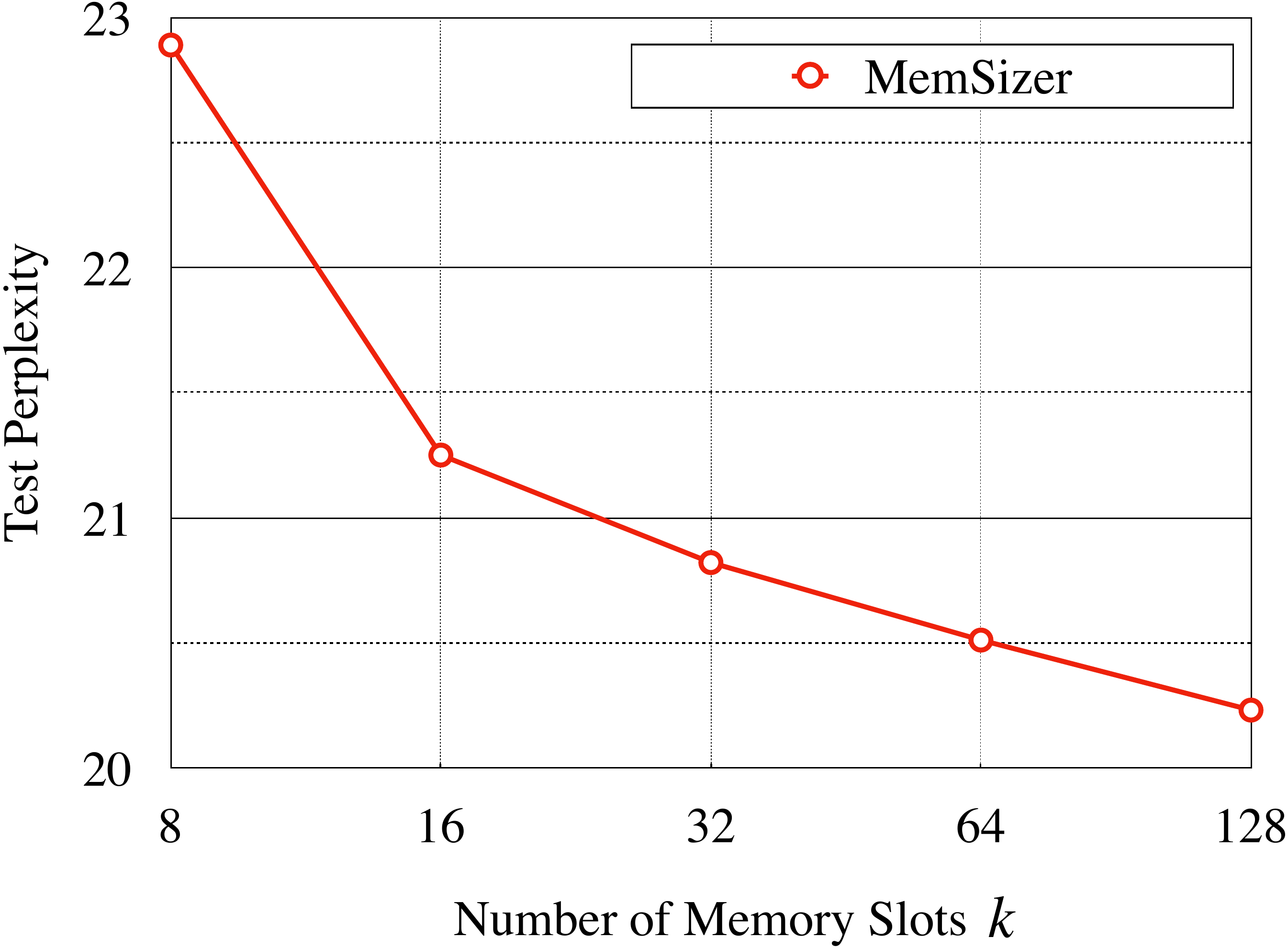}
		\caption{Effect of different numbers of memory slots $k$.}
		\label{ppl.k}
	\end{subfigure}
	\begin{subfigure}{0.49\textwidth}
		\includegraphics[scale=0.21]{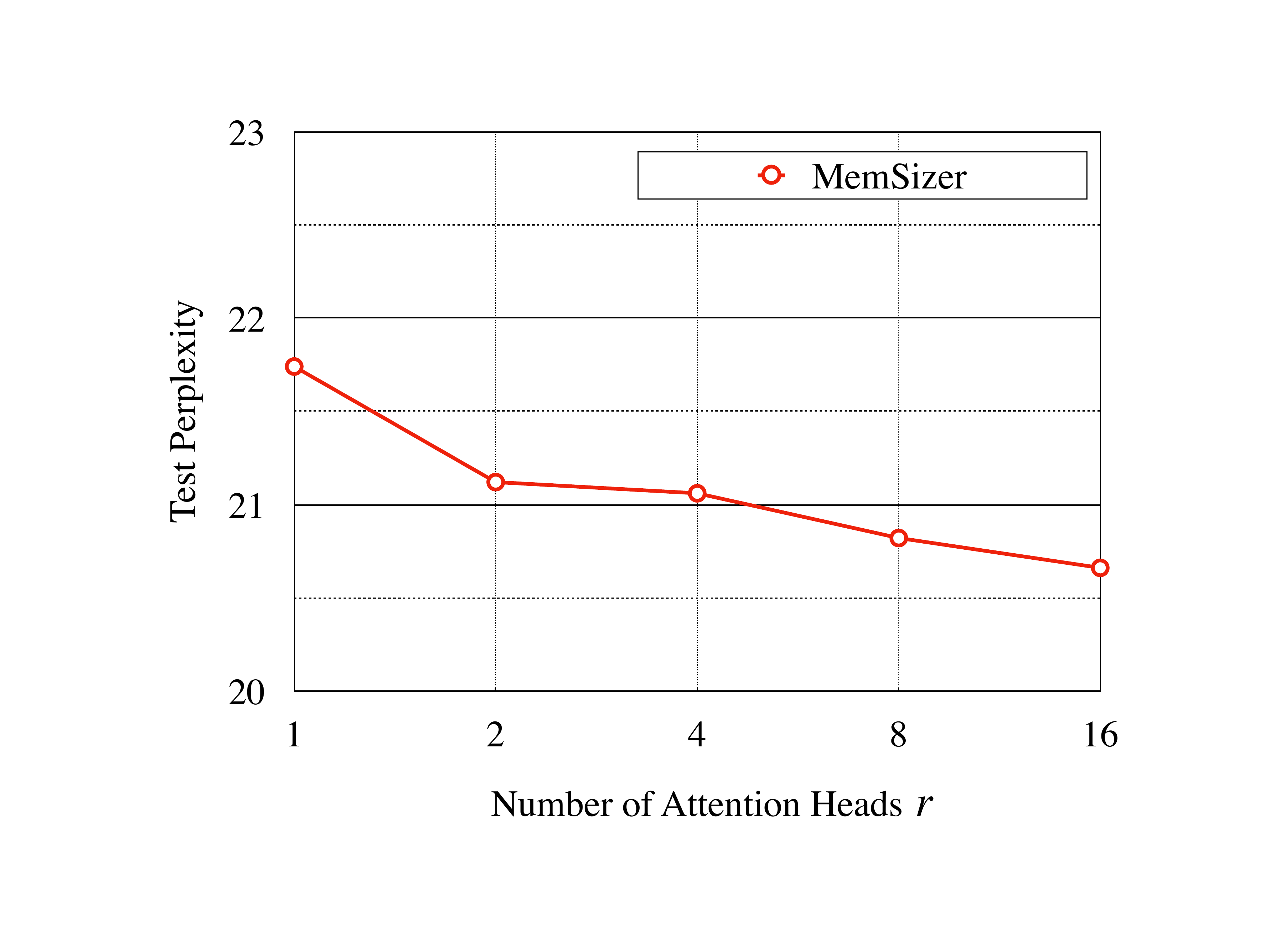}
		\caption{Effect of different numbers of attention heads $r$.}
		\label{ppl.r}
	\end{subfigure}
	\caption{Language model (Wikitext-103) perplexities of different model configurations.}
	\end{figure*}

\paragraph{Results}
Table \ref{lm_results} presents the language modeling results in perplexity and computation cost. 
We observe that \ours outperforms ELU and RFA, and achieves comparable performance to T2R, suggesting that a similar level of performance to the state-of-the-art kernel-based transformer can be obtained without approximating the softmax attention in the vanilla transformer. The generation time, memory usage, and model size are significantly reduced in \ours. We attribute this reduction to the fact that \ours: $i)$ uses fewer parameters in feature mapping as it projects the input into a much lower dimension $k$;
$ii)$ does not have the output projection layer; $iii)$ suppresses the computation of intermediate state for feature mapping required in kernel-based transformers. There remains a gap of 2.3 perplexity points between the \ours and transformer models, which might be reduced by leveraging a swap-then-finetune approach similar to \citet{kasai2021finetuning}. Further improvement of the \ours is left for future work. Compared with the results from machine translation and abstractive text summarization, we hypothesize that \ours is more advantageous with cross-attention in encoder-decoder architectures.

\subsection{Analysis of \ours}
\paragraph{Computational Overhead vs. Sequence Length}
As discussed, \ours is a linear and recurrent model for sequence generation tasks. To evaluate the time and memory efficiency against length, we run a set of experiments with different sequence lengths. For simplicity, we assume the source length is equal to the target length in our experiments \citep{kasai2021finetuning}. 
Figure \ref{time.len} and \ref{memory.len} show the time and memory cost results of MT (En-De) models in Table \ref{mt_results}. All models are
tested using greedy decoding with the same batch size of 256 on the same NVIDIA A100 GPU. As shown in figure~\ref{time.len}, we observe that \ours can generate a nearly-constant number of tokens per second regardless of the sequence length, dramatically outpacing the vanilla transformer model in longer sequence generation (300\% speedup when the length becomes 512). \ours also outperforms other linear recurrent variants by large margins (35\% faster than ELU for 512-length sequences). The maximum speedup compared with other linear recurrent variants is achieved at length=64. 
Figure~\ref{memory.len} plots decoder memory consumption when running the generation with different lengths. The curves show that the peak memory consumption is almost a constant over varying sequence lengths and is consistently lower than other baselines. This reveals the potential of \ours to achieve even more significant speed gains by allowing for a larger batch size thanks to its lower memory consumption.

\paragraph{Number of Memory Slots}
Next, we study the effect of the number of memory slots $k$. Figure \ref{ppl.k} compares the test perplexities using different values of $k$ on the WikiText-103 language model task. We observe that the performance gets better as $k$ goes larger. Among the values of $k$ in Figure \ref{ppl.k}, we do not observe that the number of memory slots $k$ has a considerable impact on inference time and memory cost. Presumably, as shown in Section~\ref{sec:memsizer}, as $k$ is generally much smaller than the model dimension $d$, a larger $k$ does not slow down the inference. However, during training time, processing time per token is roughly linear to $k$, presumably because more intermediate states need to be stored for back-propagation. 

\paragraph{Number of Attention Heads}
We also investigate the impact of the number of attention heads on model performance. Figure \ref{ppl.r} shows the results with varying values of $r$ on the WikiText-103 language model task. As can be seen, the number of attention heads slightly affects the test perplexity, resulting in slightly better performance with more attention heads. No significant difference in training and inference overhead is observed, as the multi-head computation is lightweight in \ours (\textit{e.g.}, setting $r=16$ only introduces 4.5 \% more parameters and GPU memory than $r=1$).

\paragraph{\ours with alternative design of Keys $\mK$}
We further experiment with freezing the keys $\mK$ with random standard Xavier initialization and let the input $\vq$ adapt to these keys. In both language model and machine translation tasks, the performance dropped by a relatively small margin (See Appendix~\ref{app:fixk}, Table~\ref{freeze_k}), indicating learning $\mK$ is less essential comparing to learning $\mV$. Another evidence of this is that we also performed experiments to model $\mK$ in the same input-dependent manner as $\mV$, which failed to yield performance gains.

\section{Related Work}
\label{sec:related_work}
\paragraph{Transformers with Memory Mechanism}
Previous work investigated injecting a memory mechanism into transformers. 
\citet{burtsev2020memory} augmented Transformer by adding memory tokens to store non-local representation. 
\citet{lample2019large} used a product key memory layer to substitute the feed-forward layer in Transformer.
\citet{fan2021augmenting} used a KNN-based information fetching module to enable Transformer access to external knowledge. 
Our approach is fundamentally different from them as we replace the standard attention (SA) with a key-value memory layer, which leads to linear complexity and recurrent computation.

\paragraph{Recurrent Transformers}
Previous work proposed several recurrent transformers focusing on approximating the softmax attention kernel between $\vq$ and $\vk$ by projecting them via feature map function $\vphi(\cdot)$. These recurrent variants scale at the linear time and constant space complexity in sequence length.
\citet{katharopoulos-et-al-2020} proposed $\vphi\left(\vx\right) = \elu \left(\vx\right) + 1$ and applied it to image generation.
In language modeling and machine translation tasks, RFA \citep{RFA} and Performer \citep{performer} used random features that approximate the softmax attention via Monte Carlo sampling \citep{rahimi2007, orthogonalrfa}. T2R \citep{kasai2021finetuning} used trainable feature mapping which allows smaller feature size thus further improving the efficiency. \citet{schlag2021linear} connects kernel-based transformers with previous Fast Weight Programmers.
However, approximating softmax typically needs additional steps to obtain intermediate feature mapping results. 
Instead of approximating the self-attention softmax kernel, \ours employs a key-value memory module, which suppresses these intermediate steps. The output projection step in SA is also omitted in this key-value memory module, yielding further computation and memory savings.

\paragraph{Other Efficient Transformers}
One family of efficient transformers limited the receptive fields that are attended to by sparsifying the attention patterns. Some works introduced fixed patterns of blockwise attention \citep{qiu-etal-2020-blockwise} and strided attention \citep{Child2019,longformer,bigbird}. 
\citep{sukhbaatar-etal-2019-adaptive} learned sparse attention patterns in a data-driven manner. These sparse local attention approaches reduced the computation at a cost of potentially harming the modeling capacity. 
Another family of efficient transformers compresses the context via low-rank projections to reduce memory overhead \citep{Wang2020LinformerSW, synthesizer}. Other methods add ``global tokens" as surrogates for global information exchange \citep{raeetal2020,ma2021luna} or employ clustering-based attention \citep{Kitaev2020ReformerTE,context-based-sparse,sinkhorn}. 
We compared \ours in detail with some of these efficient Transformers in the Appendix~\ref{app:comp}.

Prior work also suggested many other strategies to improve efficiency in transformers, such as factorization \citep{universal, albert}, pruning \citep{weightpruning,layerdrop}, and quantization \citep{q8bert,quantshen}.
Some of these methods present orthogonal design choices and can be integrated into our \ours model to gain further efficiency.

\section{Conclusion}
We present \ours, a method that leverages a novel key-value memory network specification to accelerate the original self-attention module. \ours compresses source information to a set of global memory entries and uses an unbalanced key-value mechanism which further leads to lightweight multi-head computation. \ours advances recent recurrent transformers with kernel approximation with lower time, memory, and storage cost during generation.
Our experiments in three standard generation tasks demonstrate that our model achieves an improved balance between efficiency and accuracy. The proposed method can be stacked with other computation reduction techniques to further advance the efficiency of transformers. 

\section*{Limitations}

This work has several limitations. 
First, there is still a performance gap between our method and the vanilla transformer in the language modeling task and CNN/Daily summarization task. We expect this can be closed by leveraging a swap-then-finetune procedure similar to \citep{kasai2021finetuning}. We left it for future work as we focus on closing the gap by training from scratch in this paper. 
It would also be interesting to make the attention sparse so that fewer memory slots are attended to further reduce the training and generation computation. We also note that the feedforward layer still takes a lot of computation, which can be further reduced by unifying the self-attention layer with the feedforward layer with memory network framework.



\section*{Broader Impact}
\label{app:bi}
This work focuses on improving the natural language processing (NLP) and general artificial intelligence (AI) research community. Our work can be leveraged to improve natural language generation (NLG) models, including but not limited to text editing, conversational agents and question answering systems. 
The \textbf{broader impact} the \textbf{risks} of this work are summarized as following:
\begin{itemize}[wide=0\parindent,noitemsep,topsep=0em]
    \item 
    This work can facilitate research in the NLG tasks in a generic manner, to potentially accelerate generations in applications like machine translation, text summarization, and virtual assistants. 
    \item This work is a fundamental research work that focuses on the technical improvement, thus we have NOT imposed additional aggressive filtering techniques to the text data we used, beyond what has been performed to the original dataset from their sources. The text data we used may have offensiveness/toxicity/fairness/bias issues that we have not been able to identify, as those are not the focus of this work. 
    \item Given the above potential risks, due to the nature of natural language generative models, we note that the generations or outputs of this work, though not likely, may reflect gender and other historical biases in the data. Under rare circumstances, the generations may exhibit a mild extent of unethical, biased, or offensive attitudes. These are known issues with current state-of-the-art text generation models. We would hope that a faster generation system like what we present can enable more iterations of further mitigation strategies for inappropriate and hallucinated generations.
    \item This work aims to advance AI technology in an environmental-friendly manner. Our proposed method can potentially reduce the carbon footprints produced by AI models.
\end{itemize}

\bibliography{all}
\bibliographystyle{acl_natbib}

\appendix

\onecolumn

\begin{center}
    {\Large \bf Appendix for Linearizing Transformer with Key-Value Memory}
\end{center}

\section{Illustration of MemSizer}
\begin{figure*}[ht!]
    \centering
    \includegraphics[scale=0.15]{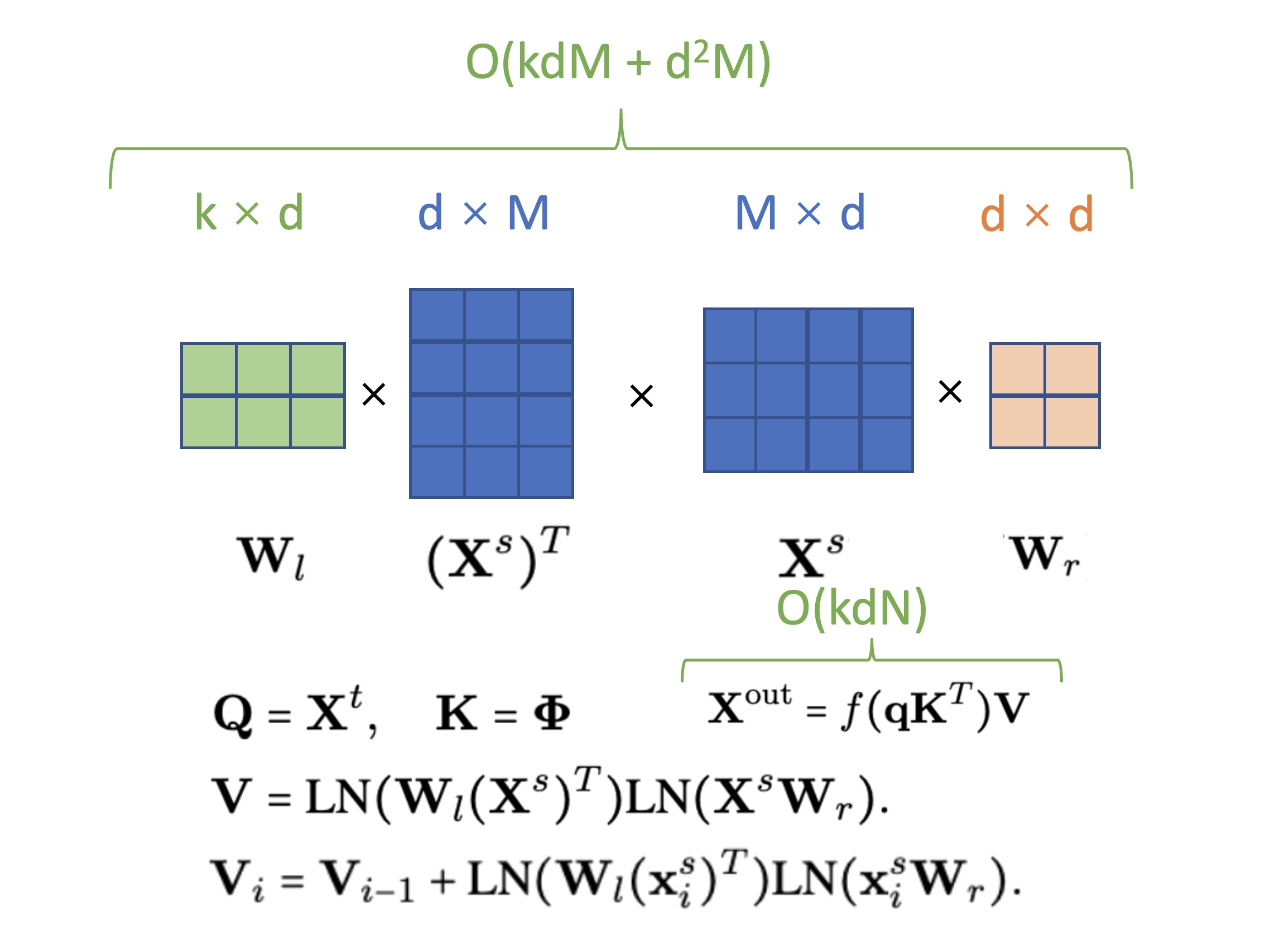}
    \caption{Illustration of the recurrent computation of Memsizer. ``LN'' represents the Layer Normalization.}
    \label{fig:v}
\end{figure*}

We provide an illustrative figure of Memsizer in Figure~\ref{fig:v}. Details are provided in the main text. 

\section{Detailed Comparison with Other Transformers}
\label{app:comp}
\begin{table*}[ht!]
\centering
\begin{tabular}{@{} lcccccc @{}}
\toprule
 & $\mQ$ & $\mK$ & $\mV$ & Linear & Recurrent & $M$-Agnostic\\
\hline
Transformer  & $\mX^{t}\mW_{q}$  & $\mX^{s}\mW_{k}$ & $\mX^{s}\mW_{v} $ &  & & \\
Dim.  & $N \times h$  & $M \times h$ & $M \times h$ & \texttimes & \texttimes& $\checkmark$\\
\hdashline
Synthesizer (R)  & $\mI$  & $\mPhi$ & $\mX^{s}\mW_{v} $ &  & & \\
Dim.  & $N \times N$  & $M \times N$ & $M \times h$ & \texttimes &\texttimes & \texttimes\\
\hdashline
Synthesizer (D)  & $\mX^{t}$  & $\mPhi$ & $\mX^{s}\mW_{v}$ &  & & \\
Dim.  & $N \times d$  & $M \times d$ & $M \times h$ & \texttimes &\texttimes & \texttimes\\
\hdashline
Linformer  & $\mX^{t}\mW_{q}$  & $\mW_{e} \mX^{s}\mW_{k}$ & $\mW_{f}\mX^{s}\mW_{v} $ &  & &\\
Dim.  & $N \times h$  & $k \times h$ & $k \times h$ & $\checkmark$ &\texttimes &\texttimes\\
\hdashline
RFA/Performer  & $\vphi(\mX^{t}\mW_{q})$  & $\vphi(\mX^{s}\mW_{k})$ & $\mX^{s}\mW_{v} $ &  & & \\
Dim.  & $N \times k$  & $M \times k$ & $M \times h$ & $\checkmark$ &$\checkmark$ &$\checkmark$\\
\hdashline
Ours  & $\mX^{t}$ & $\mPhi$ & $\mW_l (\mX^{s})^T \mX^{s} \mW_r$ &  & & \\
Dim.  & $N \times d$  & $k \times d$ & $k \times d$ & $\checkmark$ &$\checkmark$ &$\checkmark$\\
\bottomrule
\end{tabular}
\caption{A high-level comparison of attention mechanism perspectives in different transformer variants, including Synthesizer \citep{tay2021synthesizer} random/dense (R/D), Linformer \citep{Wang2020LinformerSW} and Performer \citep{performer}. 
Details are removed for brevity. 
``$M$-Agnostic'' indicates the maximum source length $M$ is \textit{not} required to be preset.
}
\label{comp}
\end{table*}
\paragraph{Comparison with Vanilla Transformer} Compared with SA in the vanilla transformer, the number of memory slots $k$ in \ours is independent of the source sequence length $M$ and can be arbitrarily configured to balance between performance and efficiency. Also, we only pack the source information $\mX^{s}$ into $\mV$.
Note that each row of $\mV$ ($\vv_{j\in \{1, \cdots, M\}}$) in the vanilla transformer corresponds to one input dimension out of the total length $M$, in a point-wise manner. However, in \ours, each memory slot value $\vv_{j\in \{1, \cdots, k\}}$ summarizes a \textit{global position-agnostic} feature of the source context $\mX^{s}$. Vanilla transformer is not linear and not recurrent. 
\paragraph{Comparison with Linformer} \ours operates with the original $\mX^{t}$ rather than the projection of $\mX^{t}$ in Linformer. The key $\mK$ in \ours does not contain source information. The projection matrices $\mW_l$ and $\mW_r$ do not depend on source dimension $M$, which allows dynamic input length thus facilitating generation. In contrast, the projection matrices $W_e$ and $W_f$ is a $k \times M$ matrix. Linformer is linear but not recurrent.

\paragraph{Comparison with Synthesizer/MLP-Mixer}
\ours also share similarities with Synthesizer \citep{tay2021synthesizer}. MLP-Mixer \citep{tolstikhin2021mlp} is computationally comparable to Synthesizer (random) except that in MLP-mixer the $f$ is an identity function.
As show in Table \ref{comp}, \ours becomes akin to Synthesizer (dense) if the $\mV$ is computed by an MLP $\mX^{s}$ ($\mV = \mX^{s}\mW_{v} + \vb_v \in \mathbb{R}^{M \times h}$). However, Synthesizer attends to $M$ different token and \ours attends on $k$ different memory slots in memory. Consequently, Synthesizer scales quadratically with input length while \ours scales linearly. 
As the maximum sequence length needs to be preset when initializing the weights, it is not straightforward to apply Synthesizer to generation tasks with various input lengths.
Synthesizer is not linear and not recurrent.

\section{\ours with fixed Keys $\mK$}
\label{app:fixk}
Inspired by the "random" version of Synthesizer \citep{synthesizer}, we further experiment with fixing the keys $\mK$ and let the input $\vq$ adapt to these keys. Specifically, we initialize $\mK$ for each layer and each head with standard Xavier initialization and freeze them during the training process. In both language model and machine translation tasks, the performance dropped by a relatively small margin (Table~\ref{freeze_k}). Presumably, as $k \ll d$, the keys in $\mK$ are almost orthogonal with Xavier initialization, thus less likely to ``collide'' with each other \citep{schlag2021linear}. Therefore, updating $\mK$ becomes less essential comparing to other parts of the model. 

\begin{table}[ht!]
\centering
\begin{tabular}{@{} lcc @{}}
\toprule
 & LM (PPL) $\downarrow$  & MT (BLEU) $\uparrow$\\
\hline
$\mK$ Trainable  & \textbf{20.8} & \textbf{28.4}\\
$\mK$ fixed  & 21.3 & 27.8 \\
\bottomrule
\end{tabular}
\caption{Fixing $\mK$ results in performance decrease. }
\label{freeze_k}
\end{table}

\end{document}